\documentclass[letterpaper, 10 pt, conference]{ieeeconf}
\IEEEoverridecommandlockouts
\usepackage{xspace}
\usepackage{colortbl} 
\usepackage{graphicx}
\usepackage{booktabs}
\usepackage{multirow} 
\usepackage[dvipsnames]{xcolor}
\usepackage{caption}
\usepackage{tabularx, microtype}
\usepackage{amsmath}
\usepackage{cuted}
\usepackage{stfloats}
\usepackage{pifont}
\usepackage{amssymb} 
\usepackage{tablefootnote}
\usepackage{float}
\usepackage{makecell}
\usepackage[normalem]{ulem}
\usepackage{bm}
\usepackage{tikz}

\usepackage{times}
\usepackage{epsfig}
\usepackage{amsfonts}
\usepackage{xcolor}
\usepackage{bbm}
\usepackage{subcaption}
\usepackage{listings}
\newcommand\copyrighttext{%
  \footnotesize \textcopyright \the\year{} IEEE. Personal use of this material is permitted. Permission from IEEE must be obtained for all other uses, including reprinting/republishing this material for advertising or promotional purposes, collecting new collected works for resale or redistribution to servers or lists, or reuse of any copyrighted component of this work in other works.}

\graphicspath{{./images/}}

\definecolor{iccvblue}{rgb}{0.21,0.49,0.74}
\newcommand{\pseudocaptionername}{LD-CPS} 
\newcommand{\pseudocaptionernamelong}{(Language-Driven Consistent PSeudo-captioner)} 
\newcommand{\pseudocaptionernamelongnobrackets}{Language-Driven Consistent PSeudo-captioner}
\newcommand{\policyname}{CLA} 
\newcommand{\largemodel}{ChatGPT}
\usepackage[pagebackref,breaklinks,colorlinks,allcolors=iccvblue]{hyperref}

\title{Embodied Image Captioning: Self-supervised Learning Agents \\ for Spatially Coherent Image Descriptions}

\author{Tommaso Galliena\textsuperscript{1,2}, Tommaso Apicella\textsuperscript{1}, Stefano Rosa\textsuperscript{1}, Pietro Morerio\textsuperscript{1}, Alessio {Del Bue}\textsuperscript{1}, Lorenzo Natale\textsuperscript{1}\\
\textsuperscript{1}Istituto Italiano di Tecnologia, Genoa, Italy\\
\textsuperscript{2}University of Genoa, Genoa, Italy \\
{\tt\small name.surname@iit.it}
}

\begin{document}
\makeatletter
\g@addto@macro\@maketitle{
  \begin{figure}[H]
  \vspace{-10pt}
  \setlength{\linewidth}{\textwidth}
  \setlength{\hsize}{\textwidth}
  \centering
  \includegraphics[width=0.99\textwidth]{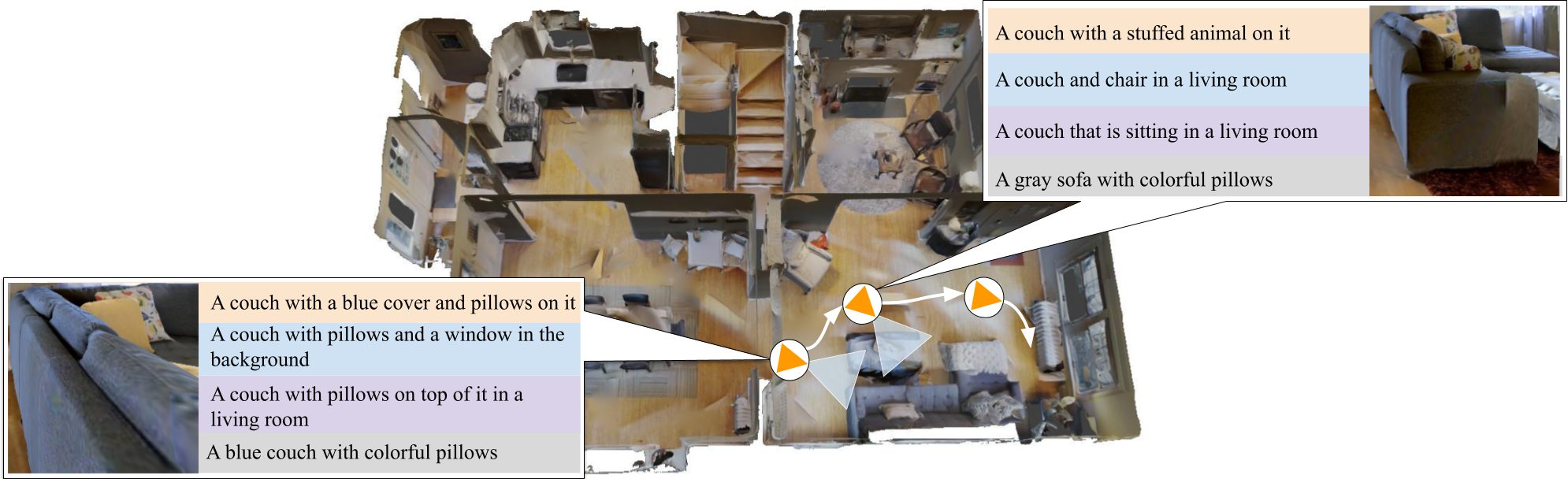}
  \caption{An agent equipped with an instance segmentation and captioning model navigates the environment. Off-the-shelf captioners predict partially wrong and inconsistent captions across different views of the same object:
\colorbox{Orange!20}{CoCa}~\cite{yu2022coca}, \colorbox{Cyan!15}{BLIP2}~\cite{li2023blip}, \colorbox{Purple!20}{Florence2}~\cite{xiao2024florence}, and \colorbox{Gray!30}{\largemodel}~\cite{openai2025chatgpt}.}
\label{fig:main_figure}
\end{figure}
}
\makeatother
\maketitle

\begin{abstract}
We present a self-supervised method to improve an agent's abilities in describing arbitrary objects
while actively exploring a generic environment. This is a challenging problem, as current models struggle to obtain coherent image captions due to different camera viewpoints and clutter. 
We propose a three-phase framework to fine-tune existing captioning models that enhances caption accuracy and consistency across views via a consensus mechanism.
First, an agent explores the environment, collecting noisy image-caption pairs. 
Then, a consistent pseudo-caption for each object instance is distilled via consensus using a large language model. 
Finally, these pseudo-captions are used to fine-tune an off-the-shelf captioning model, with the addition of contrastive learning. 
We analyse the performance of the combination of captioning models, exploration policies, pseudo-labeling methods, and fine-tuning strategies, on our manually labeled test set.
Results show that a policy can be trained to mine samples with higher disagreement compared to classical baselines. 
Our pseudo-captioning method, in combination with all policies, has a higher semantic similarity compared to other existing methods, and fine-tuning improves caption accuracy and consistency by a significant margin. Code and test set annotations available at \url{https://hsp-iit.github.io/embodied-captioning/}
\end{abstract}

\section{Introduction}
\label{sec:introduction}
Visual understanding is a fundamental skill to recognize and describe objects in a scene, enabling robots to navigate and interact with the environment~\cite{huang2023visual}. Despite advances in image captioning~\cite{yu2022coca, li2022blip, xue2023ulip, xue2024ulip, xiao2024florence}, a captioner deployed on an autonomous agent often generates wrong or inconsistent descriptions across different views of the same object, especially in the case of occlusions or challenging viewing directions (see Fig.~\ref{fig:main_figure}).

Previous works tackled the problem of inconsistent captions from three research directions: agent navigation~\cite{hong20233d, hu2023explore}, noisy captions~\cite{li2022blip, kang2023noise, jeong2024technical}, and caption comprehensiveness~\cite{chan2023ic3, sato2024caption}. Navigation-based captioning methods combine the agent observations at different times during navigation~\cite{hu2023explore}. However, when the agent moves in complex environments, the generated descriptions can be highly inconsistent for the same object. Methods that generate the most appropriate caption for an image either use the alignment between vision and language to select the description from a set of captions~\cite{jeong2024technical}, or prompt a Large Language Model (LLM) to summarize the captions~\cite{chan2023ic3}. Mentioned methods fail in case of 
inaccurate descriptions that might be selected or propagated to the captions summary. 


Compared to previous works, we tackle both the noise presence and the caption coherence as challenges inherently present when a robot navigates a complex indoor environments. To this end, we propose a three-phase framework that combines embodied perception with a self-supervised strategy to retrain the captioner enhancing the performance and increasing the consistency of generated descriptions. 
In the first phase, the agent equipped with an object detector and a captioner navigates the environment, collecting observed RGB-D images and corresponding object captions. The agent also creates a point cloud representation of the scene and clusters the object instances in 3D, given the output of the object detector. In the second phase, for each 3D segmented object instance, we prompt an LLM using the corresponding captions, their frequency and in-context learning to describe each object in the environment, hence filtering out potentially wrong captions and generating a comprehensive pseudo-label. In the third phase, we specialize the captioner through contrastive learning, maximizing the similarity of representation for different views of the same object while minimizing the similarity with other objects' representation. The framework design is modular to allow the integration of past and future approaches for exploration, pseudo-caption generation, and environment specialization, fostering fair comparisons at each phase, and offering a foundation to improve caption consistency in challenging environments.

The main contributions of this paper are the following:
\begin{itemize}
    \item A modular framework to mine agent observations during autonomous navigation, generate pseudo-caption labels, and fine-tune captioning models in a self-supervised way.
    \item A methodology to generate pseudo-captions from image samples prompting an LLM with caption frequencies and in-context learning, avoiding the time-consuming task of manual annotation of a training dataset.
    \item A fine-tuning strategy using contrastive learning to improve caption consistency across object views. 
    \item A performance analysis of the combination of exploration policies, pseudo-captioning methods and captioning models on our manually annotated test set composed of more than 8800 different images, specifically collected to assess the performance with different object views. 
\end{itemize}

\section{Related Works}
\label{sec:related}  
In this section, we review the methods in image captioning from RGB images focusing on agent navigation, noisy captions, and caption comprehensiveness to deal with inconsistent descriptions of the same object from different views.

\textbf{Image captioning.} Captioners adapt existing models by using additional inputs such as questions about the image, or bounding boxes ~\cite{xue2024ulip, li2023blip, huang2024segment, zhou2023regionblip, wang2023caption, peng2023kosmos, zhang2023gpt4roi, wang2023all, chen2023shikra} or by improving the representations using contrastive learning~\cite{yu2022coca, wang2022git, xue2023ulip, xue2024ulip, xiao2024florence, li2022blip}. To fully exploit the generalization capabilities of foundation models, the majority of methods adjusts the representation of vision-language models fine-tuning tokens or transformer layers,  reducing the gap between the pre-trained visual and text representations~\cite{li2023blip}. For example, X-InstructBLIP and BLIP2 bridge the representation between the image encoder and the text decoder by learning a set of queries to combine visual features with textual features using a lightweight transformer~\cite{xue2024ulip, li2023blip}. Other methods like SCA and CAT adapt the SAM architecture~\cite{kirillov2023segment} through visual chain-of-thoughts~\cite{wang2023caption} or additional cross-attention layers to combine the vision and text representation and predict regional captions~\cite{huang2024segment}.
A common approach among captioning methods is to learn the representation through contrastive learning i.e., maximizing the similarity among input modalities such as vision, language, and point cloud~\cite{yu2022coca, wang2022git, xue2023ulip, xue2024ulip, xiao2024florence, li2022blip}. The multimodal alignment of different object representation can be learned either during a pre-training stage enabling downstream tasks such as captioning or classification~\cite{xue2023ulip, xue2024ulip, xiao2024florence, li2022blip}, or during the training to ground the captions to the visual features~\cite{yu2022coca, wang2022git}. Despite the use of contrastive learning on input modalities, methods fail to provide consistent captions across different views of the same object, generating wrong descriptions in the case of challenging viewpoints. In this work, we apply contrastive learning to induce the model to learn close representations of the same object in the latent space while maintaining a margin distance from the representations of other objects, using triplet loss~\cite{schroff2015facenet}.

\begin{figure*}[t!]
\begin{center}\includegraphics[width=0.92\linewidth]{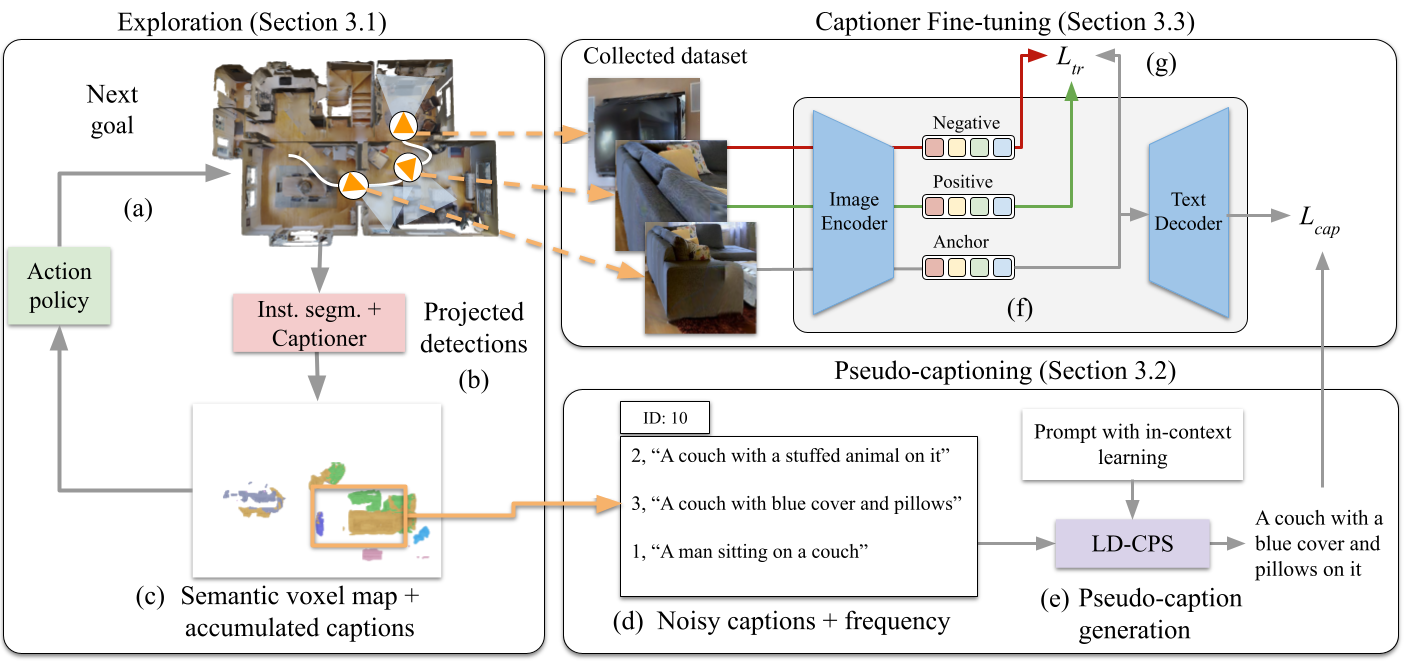}
\end{center}
\caption{Our three-phase pipeline to improve consistency of object captions. The agent navigates to a goal predicted by a policy (a). Object detections and captions are accumulated into a 3D voxel map (b). The voxel map is segmented into objects via consensus (c). Given all captions associated to an object and how many times they appear (d), \pseudocaptionernamelongnobrackets\ (\pseudocaptionername) generates a single pseudo-caption for each object (e). The captioner (f) is then fine-tuned using the pseudo-caption as supervision with the addition of a contrastive learning loss (g).}
\label{fig:framework}
\end{figure*}

\textbf{Embodied captioning.} Embodied agents navigate the environment and collect different views of the same object to learn a 3D representation of the scene~\cite{hong20233d} or to generate a caption~\cite{hu2023explore, bigazzi2021explore, bigazzi2023embodied} 
. For example, CaBOT decouples the policy that predicts the action to take in the environment, from the captioning model that generates the object description based on the trajectory~\cite{hu2023explore}. Despite generating a single caption for an object, CaBOT has prior knowledge of the best viewpoint in the scene and scenes contain few objects, limiting the application in case of challenging indoor environments like houses or office rooms and in case the annotation of the best viewpoint is not provided. Embodied scene description framework~\cite{tan2022embodied} 
used navigation to find an optimal viewpoint to describe a scene, while Bigazzi et al.~\cite{bigazzi2021explore, bigazzi2023embodied} minimized overlapping between scene descriptions. However, our objective is to improve the consistency between different views of the same object, instead of improving the variety of description.

\textbf{Noisy captions.} Training captioning models requires large-scale dataset, and collecting such data with clean annotation is a costly and time consuming task. For this reason, some methods consider noisy annotation in the datasets, either as intrinsically present~\cite{li2022blip}, as a conditioning signal~\cite{kang2023noise}, or also as a measure of what caption best represents an image~\cite{jeong2024technical}. Recently, Ensemble CLIP with consensus score (ECO) selects the most appropriate caption in a set using a linear combination between a semantic alignment between the image and captions (ensemble CLIP score) and consensus based on Term Frequency Inverse Document Frequency (TF-IDF) of n-grams~\cite{jeong2024technical}. The CLIP score represents the image-text alignment, however with partially occlusions or different viewpoints captions might contain wrong descriptions keeping the CLIP score low, hence affecting performance.

\textbf{Captions summary.} Captions describe in a comprehensive way objects in the scene. To distill a summary given different captions associated with an image, methods either prompt an LLM~\cite{chan2023ic3} or learn to maximize the variance of the caption embeddings~\cite{sato2024caption}. In particular, IC3~\cite{chan2023ic3} uses the same captioner with different temperature values to sample diverse descriptions of the same scene and an LLM prompt to summarize the caption, while PM~\cite{sato2024caption} uses different captioners to retrieve different descriptions. However, both methods consider a limited amount of noise or wrong captions that could be propagated to the final caption. 

\section{Method}
\label{sec:method}
We propose a three-phase pipeline to improve the captioning performance and the consistency of captions across different views of the same object (see Fig.~\ref{fig:framework}). In the first phase, an agent explores an environment following a policy $\pi : S_t \rightarrow a_t$, where $S_t$ is the state and $a_t$ the action at time instant $t$, and  collects RGB-D observations. 
The agent is equipped with an instance segmentation model $f_d$ and an image captioning model $f_c$. $f_d$ restricts the RGB observation $I_t \in \mathbb{R}^{W \times H \times 3}$ of the agent to detected object instances $o$, $f_d : I_t \rightarrow \{ {l}^o_t, b^o_t, M^o_t\}$, where $l \in \mathbb{R}^{O}$ is the logits vector of the $O$ classes $f_d$ can detect, $b \in \mathbb{R}^{4}$ is the object bounding box, and $M \in [0, 1]^{W \times H}$ is the object segmentation mask. $f_c$ generates the description of each detected object $f_c : I'^o_t \rightarrow c^o_t$ where $I' \in \mathbb{R}^{W^o \times H^o \times 3}$ is the object crop and $c$ the caption. 
During exploration, the agent builds a semantic voxel map of the environment consisting of $V$ points, $v_t \in \mathbb{R}^{V \times 3}$, and clusters the voxels into object IDs based on the detection logits. In the second phase, an LLM predicts a unique and concise description (pseudo-caption $\tilde{c}$) of each clustered object based on all the corresponding captions collected during exploration. In the third phase, we fine-tune the captioner using pseudo-captions to improve the consistency of captions for the same object instance. 

\subsection{Navigation and point cloud segmentation}~\label{ssec:navigation}
\begin{figure}[t!]
\centering
\includegraphics[width=\linewidth]{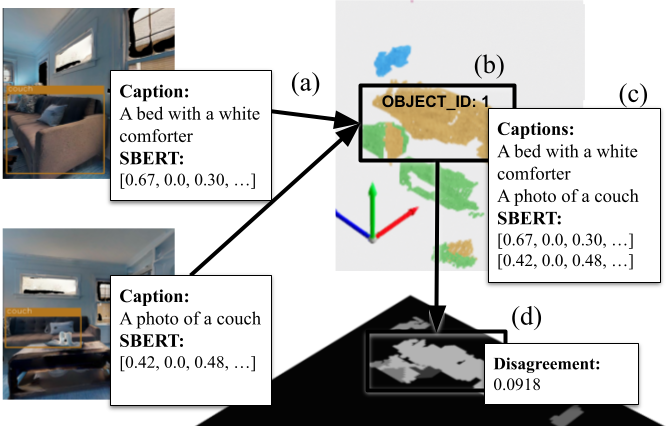}
\caption{Object clustering and disagreement computation. Object-caption pairs are projected into a voxel map (a), and  assigned to an unique object by clustering voxels according to the detected object logits (b). Captions for the same object are encoded via SBERT (c) and disagreement between encoded captions is down-projected to a 2D disagreement map (d).}
\label{fig:voxelmap-reprojection}
\end{figure}

At each time step $t$, the agent moves in the environment (Fig.~\ref{fig:framework}a) and updates the voxel map $v$ by projecting detected objects mask $M^o$ in 3D using the depth image (Fig.~\ref{fig:framework}b). We show the details of the point cloud segmentation in Fig.~\ref{fig:voxelmap-reprojection}. The agent associates to each voxel the objects logits ${l}^o$, objects mask $M^o$, and objects description $c^o$ (Fig.~\ref{fig:voxelmap-reprojection}a). For each voxel, the object pseudo-label is obtained as the \texttt{argmax} of detected logits. A 26-connectivity 3D connected components algorithm gives a unique ID (in the form of an increasing integer) to each voxel based on the pseudo-label (Fig.~\ref{fig:voxelmap-reprojection}b). All the captions associated to the voxels belonging to an object ID are grouped together.

\textbf{Learned policy (Caption-Look Around - \policyname).} The built voxel map can be used to guide an exploration policy based on maximizing the disagreement between captions of the same object (Fig.~\ref{fig:framework}a). For this reason, we adapt Look-Around~\cite{lookaround2024} policy (inspired by Active Neural Slam~\cite{chaplot2020learning}) to tackle the captioning task. 
The policy takes an input $S_t$ defined as:
\begin{equation}
S_t = \{ D^{map}_t \in [0,1]^{2 \times K \times K}, e_t \},
\end{equation}
where the first channel of $D^{map}_t$ contains a \emph{disagreement map} with canonical dimension $K\times K$, where $K=128$; the second channel contains the map of the explored environment with the agent's position superimposed; finally $e_t$ contains the global orientation of the agent with respect to the map. 
For the disagreement map, we use a per-pixel measure of inconsistency between captions, computed as follows: for each object, all captions are encoded via SBERT \cite{reimers2019sentence}.
Then, the cosine distance among all caption embeddings belonging to the same object is used as a measure of disagreement, which is down-projected onto a 2D image (see Fig.~\ref{fig:voxelmap-reprojection}d).
The policy learns to predict a long-term goal $(x,y)$ in the environment through reinforcement learning using the total disagreement as a curiosity reward~\cite{lookaround2024}.

\subsection{Language-Driven Consistent PSeudocaptioner}
The exploration phase assigns a unique ID to the object instances in the built voxel map. However, different captions $c^o_t$ are still associated with each object instance, and hence they are not consistent across views. 
We ensure consistency at the label level using a fully self-supervised approach based on prompting an LLM to perform caption distillation. We call our method \pseudocaptionername\ \pseudocaptionernamelong. 
For each object instance, \pseudocaptionername\ generates the pseudo-caption $\tilde{c}$ from the list of predicted captions $c_t$ as follows: the captions are first pre-processed to remove all uninformative text such as ``A picture of...'' or ``A photo of...'' that could be the result of bias in the captioner training or due to the distribution shift between training set and simulated environment. Then, a prompt is devised that asks an LLM to combine the list of all captions, given their frequency i.e., how many times they occur for the same object (Fig.~\ref{fig:framework}d), into a concise caption (Fig.~\ref{fig:framework}e). The prompt also highlights to the LLM the fact that some captions might be inaccurate or wrong. In addition to considering caption frequency, we use in-context learning, by providing a list of examples of how to solve the task inside the prompt.
Despite using an LLM to summarize the caption, the proposed approach is different from IC3~\cite{chan2023ic3}, since we use the frequency of the captions in the prompt, under the assumption that the noisy data are not the majority. In-context examples are also provided to improve the pseudo-caption distillation task. We show the full prompt in Sec. E of the Supp. Mat.

\subsection{Learning captions consistency}~\label{ssec:fine-tuning}
We use the pseudo-captions $\tilde{c}$ generated during the second phase as label for the captioner fine-tuning (Fig.~\ref{fig:framework}f). The captioner maps the image $I'$ to a caption $c$ supervised with the tokens (subwords) $\tilde{y}$ that make up $\tilde{c}$. Captioning models are trained and fine-tuned using the captioning loss $\mathcal{L}_{cap}$ i.e., the cross-entropy between the predicted probabilities for the tokens $p$ and the tokens $\tilde{y}$ in the pseudo-caption:

\begin{equation}
    \mathcal{L}_{cap} = - \sum_t \sum_c \tilde{y}_{t,c} \: log(p_{t,c})\:,
\end{equation}

\textbf{Triplet loss.} To improve caption consistency between different views of the same object, we add an auxiliary triplet loss $\mathcal{L}_{tr}$ (Fig.~\ref{fig:framework}g). For each image in a batch (anchor), we randomly sample a view of the same object instance (positive example) and an image of another object instance (negative example). The triplet loss pulls the visual features of a positive example $x^p$ closer to the visual features of the anchor $x^a$, while keeping a margin distance $\epsilon$ from the visual features of the negative example $x^n$: 
\begin{equation}
    \mathcal{L}_{tr} = \max\{d(x^a, x^p) - d(x^a, x^n) + \epsilon, 0\} \:
\end{equation}
where $d$ is the distance between the visual features. The final loss we minimize is a linear combination of the captioning and the triplet losses: 
\begin{equation}
    \mathcal{L} = \mathcal{L}_{cap} + \lambda_{tr} \mathcal{L}_{tr} \:,
\end{equation}
where $\lambda_{tr}$ is the coefficient weighting the triplet loss.

\section{Experimental setup}
\label{sec:experiments}
\subsection{Datasets}
We validate our framework on two datasets: Gibson~\cite{xia2018gibson} and Habitat-Matterport (HM3D)~\cite{ramakrishnan2021habitatmatterport3ddatasethm3d}. To avoid using degraded object meshes, we consider a total of 194 scenes with a reconstruction quality between 3 and 5 (the maximum), and we randomly split the scenes into training set (134), validation set (28) and test set (32) keeping the initial distribution of scene quality across the splits.  
We use the 181 scenes from the Habitat Challenge, split into a training set (126), a validation set (27), and a test set (28). We split data based on the scenes to evaluate the generalization of the framework to environments unseen during training in both Gibson and HM3D.

\subsection{Implementation details}\label{compared-methods}
\textbf{Navigation.}
We import the datasets scenes in the Habitat simulator~\cite{savva2019tplatformembodiedai} that allows agent to explore environments and to collect observations. We compare three exploration policies: \emph{random goals}, \emph{frontier exploration} and the learned policy (\emph{\policyname}) presented in Sec. \ref{ssec:navigation}. The \textit{random goals} policy chooses a feasible random goal in the map and navigates to the goal with the path planner. The \textit{frontier exploration} policy implements classical frontier-based exploration \cite{yamauchi1998frontier}: the agent keeps an internal representation of the explored map and computes feasible goals at the frontiers (the limits of the explored map), then it navigates via the path planner to the goal that provides the highest potential information gain (surrounded by the highest number of unknown map cells). We train \textit{CLA} following the Look Around setup~\cite{lookaround2024}.

\textbf{Instance segmentation.}
We equip the agent with a Mask2Former model to detect and segment objects in the environment~\cite{cheng2022masked}. 
In particular, we use Mask2Former with a Swin Transformer backbone~\cite{liu2021swintransformerhierarchicalvision}, pre-trained on COCO instance segmentation~\cite{lin2015microsoftcococommonobjects}. 
For the purposes of this work, we focus on six classes of indoor objects that are commonly found in Gibson and HM3D: couch, potted plant, bed, toilet, tv, table. The size of these categories allows to reduce Mask2Former misdetections and to obtain a sufficiently large resolution of the cropped images to be processed by a captioner. 
We increase height and width of the bounding box by 10 pixels for each side to avoid cropping the object edges and to include a part of the background (context) in the object crops. After clustering the point cloud into object instances using connected components, we render the RGB images from the collected camera poses in the environment. The bounding boxes are ray-traced from the object clusters in the point cloud to the image plane. To further limit the amount of wrong detections due to the noise in the point cloud or the presence of occlusions, we discard bounding boxes having a confidence lower than $0.7$, an area less than $8000$ pixels, and we filter them using Non-Maximum Suppression with a threshold of $0.8$ in intersection over union~\cite{1699659}. However, due to inaccuracies in the point cloud projection process, such as occlusion, depth estimation, or due to artifacts from the ray-tracing steps, the obtained bounding boxes may not align precisely with the extent of the objects in the image support. Hence, we repeat the filtering based on IoU 0.8.

\textbf{Captioning.}
We evaluate our framework using two state-of-the-art captioning models, CoCa~\cite{yu2022coca} and BLIP2~\cite{li2023blip}, selected based on their size (CoCa has almost a tenth of BLIP2 parameters).
CoCa uses contrastive learning to align images with text, while a transformer decoder combines image and text representations to generate the captions through cross-attention. We use CoCa with a ViT-L/14 backbone~\cite{radford2021learning} pretrained on LAION-2B~\cite{schuhmann2022laion} and fine-tuned on MSCOCO~\cite{chen2015microsoft}. BLIP-2 combines the visual features with the queries of a Transformer (Q-Former), and generates the caption processing the output of the Q-Former with an LLM. We use BLIP2 quantized to half precision and pretrained on LAION~\cite{schuhmann2022laion}, with a CLIP image encoder~\cite{radford2021learning} and an OPT-2.7b LLM~\cite{zhang2022opt}. 

\textbf{Pseudo-captioning.} We compare our method for pseudo-caption generation with two methods in the literature: ECO~\cite{jeong2024technical}, and IC3~\cite{chan2023ic3}. ECO selects the most appropriate caption from a set of captions~\cite{jeong2024technical}, combining a similarity score computed using an ensemble of CLIP-based models~\cite{radford2021learning} with a consensus score derived from CIDEr~\cite{vedantam2015cider}. We take as a pseudo-caption the description that shows the highest ECO score. IC3 first generates different captions for the same image using temperature-based sampling. Then an LLM is prompted to discard potentially incorrect captions and to distill a single more detailed description from the set of captions~\cite{chan2023ic3}. We keep the same prompt to summarize the captions in a single description (pseudo-caption) starting from the different captions for each object instance. We use LLaMa-3~\cite{dubey2024llama3herdmodels} as LLM for our pseudo-captioning method and IC3.

\textbf{Training and testing images.} We launch an exploration episode of 300 steps for each environment in both Gibson and HM3D. Each episode starts from a random position. 
We equip the agent with CoCa to keep the computational burden of the navigation policy low, as CoCa is 4 times faster than BLIP2. 
Since each policy explores the testing environments in a different way, we merge the different sets collected with each policy into a single test set. For each object instance in the test set, we manually annotated a single, concise, and representative caption, by looking at multiple images from different viewpoints. 
We associate these new annotations with the corresponding object observations collected during agent navigation leveraging the point cloud segmentation of the scene. 

\textbf{Learning captions consistency.} We compare two fine-tuning strategies for captioners: standard fine-tuning that minimizes only the captioning loss $\mathcal{L}_{cap}$, and an alternative approach that uses contrastive learning with a triplet loss. We evaluate both approaches against the off-the-shelf models to assess their impact on captioning performance and consistency.
We initialize CoCa and BLIP-2 with off-the-shelf weights and we fine-tune them for 10 epochs, applying early stopping if the validation loss does not decrease for 3 consecutive epochs. In the triplet loss, we set the weight $\lambda_{tr}$ to $0.1$, the margin $\epsilon$ to 2, and $d$ as the Euclidean distance. For CoCa, we disable the contrastive loss to prevent penalizing the encoders based on vision-language alignment. Given BLIP-2’s large number of parameters, we apply LoRa adaptation~\cite{hu2022lora} on the query and value projections of the Q-Former module 
using the Rank-Stabilized version~\cite{kalajdzievski2023rank}. A complete list of hyperparameters is provided in Sec. A of the Supp. Mat. Additionally, we compare also with the off-the-shelf version of ChatGPT (o1)~\cite{openai2025chatgpt} and Florence2~\cite{xiao2024florence}, fed with the object crops. We use Florence2-base in image captioning mode, and we prompt ChatGPT to use no more than 5-7 words, only one adjective per-noun, and we provide some examples of sentences. The full prompt is in Sec. E of the Supp. Mat.  

\subsection{Evaluation metrics}~\label{ssec:performance-measures}
We compare the performance of methods using the following measures: BLEU-4 ($B_4$) measures the average precision up to 4-grams (4 adjacent words) i.e., percentage of predicted 4-grams occurring in the annotation~\cite{papineni2002bleu}; 
ROUGE-L ($R_L$) computes the longest sequence of words shared between prediction and annotation (LCS), and combines the percentage of predicted words in the LCS (precision) with the percentage of annotation words in the LCS (recall) through a harmonic mean~\cite{lin2004rouge}; 
METEOR ($M$) compares 1-grams of predicted and annotated captions by matching exact words, stems, synonyms, and paraphrases. $M$ combines precision and recall using a harmonic mean, adding a penalty for fragmented sentences to encourage fluency and coherence;
CIDEr ($CI$) computes the average cosine similarity between the n-gram TF-IDF in the predicted caption and in the annotation~\cite{vedantam2015cider}; SPICE ($SP$) compares the semantic content of the elements in the predicted and annotated captions (nouns, attributes and their relationship represented as scene graph) by computing the harmonic mean between precision (percentage of predicted elements occurring in the annotation) and recall (percentage of annotated elements occurring in the prediction); 
the cosine similarity ($CS$) between SBERT embeddings considers the semantic similarity between the predicted and annotated captions~\cite{reimers2019sentence}. We follow available performance measures formulation and implementation~\cite{chen2015microsoft, reimers2019sentence}.

\section{Results}
\subsection{Exploration policy}
\begin{figure}[t]
\begin{center}
    \includegraphics[width=\linewidth]{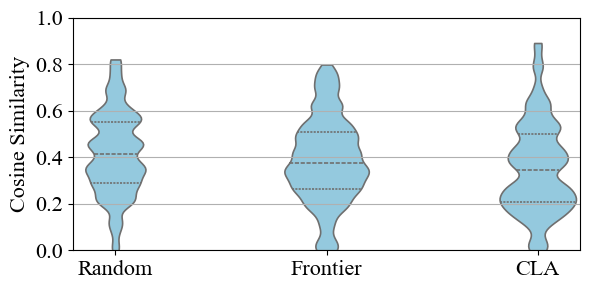}
\end{center}
\caption{Distribution of cosine
similarity between SBERT embeddings of CoCa captions across object instances in Gibson test set, for each exploration policy.}
\label{fig:policy_disagreement_distribution}
\end{figure}
We first compare the exploration policies in terms of cosine similarity between the SBERT embeddings of all CoCa captions assigned to each object instance in the corresponding test set (the lower the cosine similarity the better because the descriptions of the same object are semantically different). In Fig.~\ref{fig:policy_disagreement_distribution} we show the distribution of cosine similarity values in the Gibson test set using CoCa as a captioner. For all policies, 75\% of the data shows a cosine similarity lower than 0.6. Random gives the highest similarity at the first, second, and third quartile compared to frontier and CLA. CLA mines 50\% of data (median) with similarity values lower than other policies, and the highest density value (number of cosine similarity samples) is almost in correspondence first quartile, while for random and frontier between the first and the second quartile. Note that CLA shows the highest value of cosine similarity due to the fact that the policy can initially visit also objects with consistent captions, in order to build the disagreement map. 

\begin{table}[t!]
    \centering
    \scriptsize
    \setlength\tabcolsep{1.6pt}
    \renewcommand{\arraystretch}{0.9}
    \caption{Comparison of pseudo-captioning performance on Gibson and HM3D testing sets. In \textbf{bold} the highest values among pseudo-captioning methods, in  \colorbox{Blue!20}{blue} among the policies.}
    \begin{tabular}{r r r r c c c c c c}
    \toprule
     & \textbf{Captioner} & \textbf{Policy} & \textbf{Pseudo-caption} & $B_4$ & $M$ & $R_L$ & $CI$ & $SP$ & $CS$ \\
    \midrule
    \multirow{18}{*}[-10pt]{\rotatebox[origin=c]{90}{Gibson}} & \multirow{9}{*}[-5pt]{CoCa} & \multirow{3}{*}{Random} & ECO & 10.07 & 22.92 & 43.29 & 0.76 & 27.29 & 69.43\\
    & & & IC3 & 1.25 & 12.73 & 14.51 & 0.02 & 10.58 & 56.68 \\\
    & & & \pseudocaptionername & \textbf{14.70} & \textbf{25.13} & \textbf{50.15} & \textbf{1.19} & \textbf{32.16} & \textbf{71.71} \\
    \cmidrule{3-10}
    &  & \multirow{3}{*}{Frontier} & ECO & 10.29 & 20.66 & 44.36 & 0.79 & 32.56 & 70.50\\
     &  &  & IC3 & 1.59 & 14.06 & 15.15 & 0.00 & 11.29 & 58.32 \\
     &  &  & \pseudocaptionername & \textbf{12.38} & \textbf{21.93} & \textbf{47.51} & \textbf{0.90} & \textbf{33.14} & \textbf{70.80} \\
    \cmidrule{3-10}
    &  & \multirow{3}{*}{CLA} & ECO & 15.84 & \cellcolor{blue!20}\textbf{26.34} & 46.08 & 1.18 & 30.91 & 74.28 \\
     &  &  & IC3 & 2.67 & 15.57 & 17.45 & 0.00 & 12.74 & 60.34 \\
     &  &  & \pseudocaptionername & \cellcolor{blue!20}\textbf{21.05} & 25.82 & \cellcolor{blue!20}\textbf{52.63} & \cellcolor{blue!20}\textbf{1.54} & \cellcolor{blue!20}\textbf{34.52} & \cellcolor{blue!20}\textbf{75.01} \\
    \cmidrule{2-10}
    & \multirow{9}{*}[-5pt]{BLIP2} & \multirow{3}{*}{Random} & ECO & 11.82 & 18.32 & 31.04 & 0.23 & 19.60 & 61.09 \\
    & & & IC3 & 1.44 & 13.02 & 15.2 & 0.00 & 9.39 & 56.7 \\
    & & & \pseudocaptionername & \textbf{15.49} & \textbf{23.43} & \textbf{49.42} & \textbf{1.07} & \textbf{30.18} & \textbf{67.34} \\ 
    \cmidrule{3-10}
    & & \multirow{3}{*}{Frontier} & ECO & 12.04 & \textbf{24.94} & 48.81 & 1.10 & \cellcolor{blue!20}\textbf{38.10} & \textbf{73.17} \\
     &  &  & IC3 & 2.10 & 15.33 & 17.41 & 0.00 & 11.91 & 57.00 \\
     &  &  & \pseudocaptionername & \textbf{16.67} & 23.66 & \textbf{50.36} & \textbf{1.51} & 35.41 & 70.58 \\
    \cmidrule{3-10}
    & & \multirow{3}{*}{CLA} & ECO & 15.99 & 25.49 & 45.59 & 1.20 & 30.21 & 74.79 \\
     & & & IC3 & 2.65 & 15.30 & 16.09 & 0.00 & 12.18 & 60.68 \\
     & & & \pseudocaptionername & \cellcolor{blue!20}\textbf{21.30} & \cellcolor{blue!20}\textbf{25.67} & \cellcolor{blue!20}\textbf{51.54} & \cellcolor{blue!20}\textbf{1.56} & \textbf{34.28} & \cellcolor{blue!20}\textbf{74.94} \\
     
    \midrule
    \multirow{18}{*}[-10pt]{\rotatebox[origin=c]{90}{HM3D}} & \multirow{9}{*}[-5pt]{Coca} & \multirow{3}{*}{Random} & ECO & 15.04 & 25.04 & 46.20 & 1.06 & 30.54 & 73.33 \\
     & & & IC3 & 2.37 & 15.52 & 17.09 & 0.00 & 11.41 & 59.54 \\
     & & & \pseudocaptionername & \textbf{19.56} & \cellcolor{blue!20}\textbf{25.56} & \textbf{52.59} & \textbf{1.38} & \textbf{32.06} & \textbf{74.81} \\
    \cmidrule{3-10}
    & & \multirow{3}{*}{Frontier} & ECO & 17.79 & \textbf{23.63} & 48.83 & 1.27 & 32.01 & 76.41 \\
     &  &  & IC3 &1.68 & 12.80 & 13.49 & 0 & 10.49 & 48.97 \\
     &  &  & \pseudocaptionername & \textbf{19.85} & 23.57 & \cellcolor{blue!20}\textbf{52.72} & \cellcolor{blue!20}\textbf{1.57} & \cellcolor{blue!20}\textbf{37.49} & \cellcolor{blue!20}\textbf{78.41} \\
    \cmidrule{3-10}
     & & \multirow{3}{*}{CLA} & ECO & 17.79 & \textbf{24.69} & 46.19 & 1.29 & 33.34 & \textbf{74.36} \\
     &  &  & IC3 & 1.72 & 12.66 & 13.20 & 0.00 & 9.81 & 46.65 \\
     &  &  & \pseudocaptionername & \cellcolor{blue!20}\textbf{20.63} & 23.19 & \textbf{49.47} & \textbf{1.36} & \textbf{35.19} & 73.95 \\
     \cmidrule{2-10}
    & \multirow{9}{*}[-5pt]{BLIP2} & \multirow{3}{*}{Random} & ECO & 14.32 & 24.66 & 48.71 & 1.06 & 28.44 & \textbf{73.28}\\
     &  &  & IC3 & 2.44 & 14.87 & 16.62 & 0.00 & 10.10 & 56.85 \\
     &  &  & \pseudocaptionername & \textbf{19.20} & \cellcolor{blue!20} \textbf{24.91} & \textbf{51.41} & \textbf{1.27} & \textbf{29.64} & 70.18 \\
    \cmidrule{3-10}
     & & \multirow{3}{*}{Frontier} & ECO & 17.78 & 20.09 & 44.98 & 1.13 & 23.46 & 63.40 \\
     &  &  & IC3 & 3.24 & 12.81 & 14.04 & 0.09 & 10.1 & 19.26 \\
     &  &  & \pseudocaptionername & \textbf{17.48} & \textbf{22.51} & \textbf{48.87} & \textbf{1.25} & \textbf{26.69} & \textbf{68.50} \\
     \cmidrule{3-10}
     & & \multirow{3}{*}{CLA} & ECO & 15.60 & 23.00 & 45.83 & 1.06 & 33.29 & 72.37 \\
     &  &  & IC3 & 1.43 & 11.80 & 12.78 & 0.00 & 8.37 & 44.23 \\
     &  &  & \pseudocaptionername & \cellcolor{blue!20}\textbf{22.75} & \textbf{24.48} & \cellcolor{blue!20}\textbf{53.06} & \cellcolor{blue!20}\textbf{1.62} &\cellcolor{blue!20} \textbf{35.58} & \cellcolor{blue!20}\textbf{74.09} \\
    \bottomrule
    \addlinespace[\belowrulesep]
    \multicolumn{10}{l}{\parbox{0.98\linewidth}{\scriptsize{KEYS -- $B_4$: BLEU, $M$: METEOR, $R_L$: ROUGE-L, $CI$: CIDEr, $SP$: SPICE, $CS$: cosine similarity between SBERT embedding of prediction and annotation.}}}
    \end{tabular}
    \label{tab:pseudo_caption_results}
\end{table}

\subsection{Pseudo-captioning}
We evaluate the performance of pseudo-caption generation using the metrics described in Sec.~\ref{ssec:performance-measures} and we compare our method with two existing approaches, ECO~\cite{jeong2024technical} and IC3~\cite{chan2023ic3} (described in Sec.~\ref{compared-methods}) on the corresponding test sets of Gibson and HM3D manually annotated. Results in Table~\ref{tab:pseudo_caption_results} show that in general our method outperforms both ECO and IC3 across nearly all evaluation metrics in both Gibson and HM3D, for each captioner and policy (values in bold). IC3 has always lower values compared to other methods, due to the fact that the pseudo-caption is a summary of the initial set of captions, hence inaccurate details propagate to the eventual description. Moreover, ECO depends on CLIP models that might select the wrong pseudo-caption to associate with the image. Our method shows higher $CI$ values, despite ECO uses CIDEr-derived consensus score. In Gibson, the highest improvements of our method compared to ECO can be observed when using CoCa, in the semantic structure of captions ($SP$) with +4.87 percentage points (p.p.) using Random policy and +4 p.p. when using CLA. A similar trend can be observed in HM3D for both CoCa and BLIP2.
Table~\ref{tab:pseudo_caption_results} also shows a comparison of the results based on the policy (values in gray). In particular, mining samples with CLA leads in most cases to higher pseudo-captioning performance for ECO and for our method compared to Random and Frontier policies, thanks to the maximization of the disagreement map. Using CLA, the improvement in similarity between predicted and annotated captions ($CS$) for CoCa in Gibson is more than 3 p.p., and for BLIP2 in HM3D the improvement in terms of SPICE ($SP$) is almost 6 p.p.

\begin{figure}[t!]
\begin{center}
    \includegraphics[width=\linewidth]{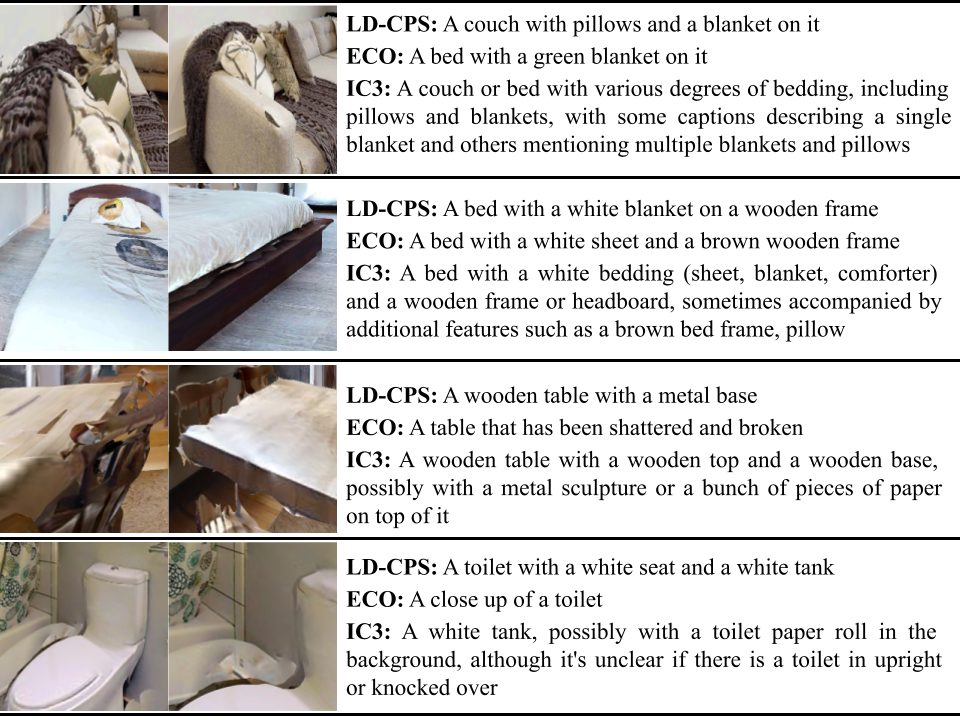}
\end{center}
\caption{Qualitative examples of pseudo-captions generated by ECO~\cite{jeong2024technical}, IC3~\cite{chan2023ic3} and LD-CPS (ours), on HM3D (first and second rows) and Gibson (third and fourth rows).}
\label{fig:pseudo_captions_qualitative}
\end{figure}

Fig.~\ref{fig:pseudo_captions_qualitative} shows examples of pseudo-captions predicted by our method, IC3 and ECO. IC3 overlooks that the input captions may be incorrect, propagating the errors in the caption summary. ECO partly relies on CLIP score, which can give inaccurate image-caption similarity scores for cluttered or partially occluded objects, thus selecting a noisy description. In contrast, our approach mitigates these issues through the information about caption frequency to discard noisy captions with low frequency, and in-context learning to provide examples of concise captions. 

\subsection{Captions consistency}
Since \pseudocaptionername\ outperforms the other methods in almost all metrics, we select it as the pseudo-captioning method to fine-tune the captioner. Table~\ref{tab:finetuning_testing_set_performance} compares the performance of different training procedures and off-the-shelf captioners on the Gibson and HM3D test sets, varying the exploration policy. In general, ChatGPT outperforms all the off-the-shelf captioners, yet the captioning performance of CoCa and BLIP2 improves after fine-tuning, with values also higher than the general purpose model ChatGPT, thanks to the learning of a unique and concise caption generated by our framework. With the majority of combinations of exploration policy and captioner, the use of triplet loss, which forces predictions of different object views to be similar, improves the captioning scores compared to standard fine-tuning in both Gibson and HM3D. 
We show an ablation study of triplet loss weight in Sec. B and qualitative comparisons of the predictions before and after fine-tuning in Sec. D of Supp. Mat.
\begin{table}[t!] 
    \centering
    \scriptsize
    \setlength\tabcolsep{2pt}
    \resizebox{\columnwidth}{!}{
    \begin{tabular}{r r r r r c c c c c c}
    \toprule
     & \textbf{Captioner} & \textbf{Pseudo-caption} & \textbf{Policy}  & \textbf{Fine-tuning} & $B_4$ & $M$ & $R_L$ & $CI$ & $SP$ & $CS$ \\
    \midrule
    \multirow{17}{*}[-20pt]{\rotatebox[origin=c]{90}{Gibson}} & Florence2 & - & - & - & 10.27 & 20.01 & 38.82 & 0.52 & 24.07 & 61.61
 \\
    & \largemodel & - & - & - & 14.44 & 22.02 & 50.61 & 1.09 & 36.26 & 70.85 \\
    \cmidrule{2-11}
    & \multirow{7}{*}[-7pt]{CoCa} & - & - & NoFT & 12.15 & 21.47 & 42.83 & 0.63 & 28.30 & 63.30 \\
    & & \multirow{6}{*}{\pseudocaptionername} & \multirow{2}{*}{Random} & Standard & 19.55 & 23.36 & 51.10 & 1.01 & 33.21 & 70.11\\
    & & & & Triplet & \textbf{19.85} & \textbf{24.83} & \textbf{53.10} & \textbf{1.12} & \textbf{35.58} & \textbf{73.26} \\
    \cmidrule{4-11}
    & &  & \multirow{2}{*}{Frontier} & Standard & \textbf{17.93} & \textbf{24.92} & 51.00 & \textbf{1.08} & 35.65 & 73.20 \\
    & & & & Triplet & 17.17 & 24.82 & \textbf{51.43} & 0.94 & \textbf{36.10} & \textbf{73.44} \\
    \cmidrule{4-11}
    & & & \multirow{2}{*}{CLA} & Standard &  13.70 & 20.50 & 45.43 & \textbf{0.72} & \textbf{26.48} & 61.76 \\
    & & & & Triplet & \textbf{14.96} & \textbf{20.79} & \textbf{45.50} & 0.71 & 25.30 & \textbf{63.21} \\
    \cmidrule{2-11}
    & \multirow{7}{*}{BLIP2} & - & - & NoFT & 11.50 & 22.64 & 43.16 & 0.58 & 26.67 & 64.48 \\
    & & \multirow{6}{*}{\pseudocaptionername} & \multirow{2}{*}{Random} & Standard & 10.27 & 23.19 & 45.51 & 0.73 & \textbf{30.78} & 66.59 \\
    & &  &  & Triplet & \textbf{12.09} & \textbf{23.39} & \textbf{46.15} & \textbf{0.75} & 29.80 & \textbf{67.43} \\
    \cmidrule{4-11}
    & &  & \multirow{2}{*}{Frontier} & Standard & 12.56 & 23.20 & 46.25& 0.71 & 30.35 & \textbf{69.75}\\
    & &  &  & Triplet &  \textbf{12.86} & \textbf{23.26} & \textbf{46.51} & \textbf{0.74} & \textbf{30.61} & 68.67\\
    \cmidrule{4-11}
    & &  & \multirow{2}{*}{CLA} & Standard & 13.35 & 23.39 & 47.82 & 0.82 & 31.59 & \textbf{69.19}\\
    & &  &  & Triplet &  \textbf{14.12} & \textbf{23.69} & \textbf{47.92} & \textbf{0.87} & \textbf{32.12} & 68.56\\
    \midrule
    \multirow{17}{*}[-20pt]{\rotatebox[origin=c]{90}{HM3D}} & Florence2 & - & - & - &  08.69 & 15.99 & 34.98 & 0.31 & 18.84 & 58.43 \\
    & \largemodel & - & - & - & 9.44 & 17.68 & 44.79 & 0.85 & 29.59 & 68.41 \\
    \cmidrule{2-11}
    & \multirow{7}{*}{CoCa} & - & - & NoFT & 09.93 & 17.36 & 38.91 & 0.44 & 22.19 & 62.08 \\
    & & \multirow{6}{*}{\pseudocaptionername} & \multirow{2}{*}{Random} & Standard & \textbf{17.34} & \textbf{21.68} & \textbf{47.46} & 0.82 & \textbf{26.90} & 72.67 \\
    & &  &  & Triplet & 16.57 & 21.39 & 46.73 & \textbf{0.83} & 29.36 & \textbf{72.77} \\
    \cmidrule{4-11}
    & &  & \multirow{2}{*}{Frontier} & Standard &  19.04 & 21.73 & 47.12 & 0.87 & 28.64 & 72.47 \\
    & &  &  & Triplet & \textbf{19.11} & \textbf{22.05} & \textbf{49.02} & \textbf{0.95} & \textbf{30.21} & \textbf{73.99} \\
    \cmidrule{4-11}
    & & & \multirow{2}{*}{CLA} & Standard &  16.35 & 20.62 & 46.19 & 0.80 & 28.38 & 70.19\\
    & &  &  & Triplet & \textbf{17.57} & \textbf{20.73} & \textbf{47.81} & \textbf{0.81} & \textbf{29.36} & \textbf{72.23} \\
    \cmidrule{2-11}
    & \multirow{7}{*}{BLIP2} & - & - & NoFT & 10.14 & 17.71 & 40.06 & 0.49 & 23.03 & 64.26 \\
    & & \multirow{6}{*}{\pseudocaptionername} & \multirow{2}{*}{Random} & Standard & 15.43 & 20.90 & 46.82 & 0.78 & 27.18 & 69.39 \\
    & &  &  & Triplet &   \textbf{15.59} & \textbf{24.14} & \textbf{49.17} & \textbf{0.91} & \textbf{32.15} & \textbf{69.47}\\
    \cmidrule{4-11}
    & &  & \multirow{2}{*}{Frontier} & Standard & \textbf{14.12} & 20.39 & 46.12 & 0.68 & 25.98 & \textbf{71.36} \\
    & &  &  & Triplet &  12.92 & \textbf{22.73} & \textbf{46.63} & \textbf{0.70} & \textbf{29.75} & 68.98 \\
    \cmidrule{4-11}
    & &  & \multirow{2}{*}{CLA} & Standard & \textbf{14.71} & 20.42 & 46.98 & 0.67 & 26.70 & 70.93 \\
    & &  &  & Triplet & 12.84 & \textbf{24.92} & \textbf{47.30} & \textbf{0.77} & \textbf{30.81} & \textbf{71.02} \\
    \bottomrule
    \end{tabular}
    }
    \caption{Comparison of captioning performance on Gibson and HM3D testing sets, varying fine-tuning method and policy. In \textbf{bold} the highest values between fine-tuning approaches. 
    KEYS -- $B_4$: BLEU, $M$: METEOR, $R_L$: ROUGE-L, $CI$: CIDER, $SP$: SPICE, $CS$: cosine similarity between SBERT embedding of prediction and annotation, NoFT: no fine-tuning.}
    \label{tab:finetuning_testing_set_performance}
\end{table}

\begin{figure}[t!]
    \centering
    \begin{subfigure}[b]{0.45\textwidth}
         \centering
         \includegraphics[width=\textwidth]{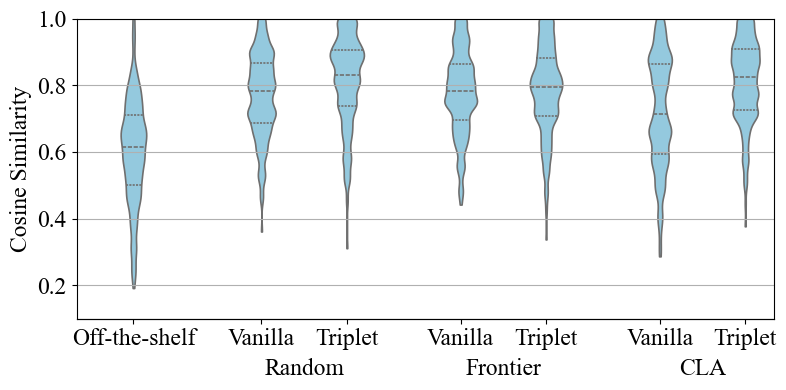}
         \caption{Gibson test set}
         \label{fig:y equals x}
     \end{subfigure}
     \vfill
     \begin{subfigure}[b]{0.45\textwidth}
         \centering
         \includegraphics[width=\textwidth]{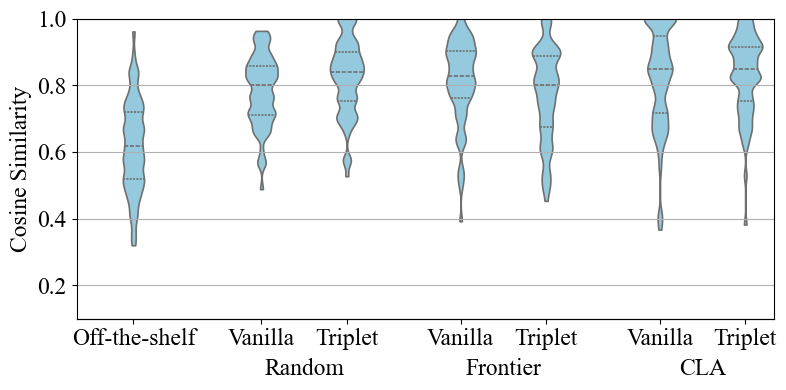}
         \caption{HM3D test set}
         \label{fig:three sin x}
     \end{subfigure}
\caption{Distribution of CoCa captions consistency across different views of the same object pre vs. post fine-tuning in Gibson and HM3D test sets.}
\label{fig:captions_consistency}
\end{figure}

Fig.~\ref{fig:captions_consistency} shows the distributions of descriptions consistency (cosine similarity among SBERT embeddings of captions belonging to the same object instance) in the test set, comparing the off-the-shelf version of CoCa with the fine-tuned one in Gibson and HM3D. In general, after the vanilla fine-tuning the distributions show an increased cosine similarity of object captions compared to the off-the-shelf model. In Gibson, all the quartiles have values higher than the vanilla fine-tuning regardless of the exploration policy, while in HM3D Random and CLA show an improvement. In the vast majority of cases, the triplet penalization improves the intra-class captions consistency.  

\section{Conclusion}\label{sec:conclusion}
We introduced our modular framework to improve the consistency of captions across different object viewpoints, based on three separate phases: navigation, pseudo-caption generation, and fine-tuning. 
The experimental validation shows that: the navigation policy explores regions with semantic disagreement among object descriptions higher than baselines; the proposed pseudo-captioning method based on captions frequency and in-context learning, along with the learned policy, generates concise captions having semantic similarity with the human annotation higher than baselines; the triplet loss fine-tuning specializes the caption descriptions to the explored environments while increasing the caption consistency of object instances. 

\textbf{Limitations.} Current limitations of the framework lay in generating wrong pseudo-captions in case the initial captions set contains a majority of repeated inaccurate descriptions, due to the proposed prompt based on caption frequency, and in aggregating inaccurate object instances in the point cloud due to the assumption of accurate camera pose and depth estimation during the point cloud generation and segmentation. Moreover, the framework is not optimized to run real-time on a standalone robot. 
Future works will focus on extending our approach to dynamic scenarios and on validating the framework on a real robot.


\section*{Acknowledgement}
We acknowledge the financial support from PNRR MUR Project PE0000013 "Future Artificial Intelligence Research (hereafter FAIR)", funded by the European Union – NextGenerationEU, CUP J53C22003010006.
This project has received funding from the European Union's Horizon research and innovation programme G.A. n. 101070227 (CONVINCE).

\bibliographystyle{IEEEtran}
\bibliography{IEEEabrv,references}

\appendices
\section{Experimental setup}
\textbf{Datasets.}
Gibson reflects the complexity of 572 real-world environments featuring object categories such as tables, sofas, plants, in different settings (rooms and object instances). HM3D comprises 1,000 reconstructed indoor residential and commercial environments having higher quality than Gibson~\cite{ramakrishnan2021habitatmatterport3ddatasethm3d}. 

\textbf{Methods under comparison.}
We compare three exploration policies: \emph{random goals}, \emph{frontier exploration} and the learned policy (\emph{\policyname}). The first two policies use the same path planner that first converts the explored map into a visibility graph (nodes represent feasible goals and edges the straight trajectory without obstacles) via skeletonization, then computes the optimal trajectory between the current position and the goal using graph search, considering nodes in the trajectory as sub-goals. Moreover, we delete repeated samples caused by the agent revisiting certain poses during exploration, to train and test on different images. Training images are augmented with 50\% probability using horizontal flip, gaussian noise and affine transformations.

\textbf{Captioners fine-tuning setup.}
We report the training setup used to fine-tune CoCa~\cite{yu2022coca} and BLIP2~\cite{li2023blip} in Table~\ref{tab:finetuning_setup}. Note that we use LoRA~\cite{hu2022lora} to fine-tune BLIP2 due to the large amount of parameters to reduce instability during the fine-tuning processing and reduce the amount of time needed for the fine-tuning processing.
\begin{table}[t!]
    \centering
    \label{tab:hyperparams}
    \begin{tabular}{r c c}
        \toprule
        Hyperparameter & CoCa & BLIP2 \\
        \midrule
        Learning Rate & 0.0005 & 0.0001\\
        lr scheduler & cosine & - \\
        Batch Size & 64 & 64 \\
        Num Workers & 4 & 4 \\
        Optimizer & AdamW & AdamW \\
        Weight Decay & 0.001 & 0.001 \\
        Epochs & 10 & 10 \\
        Patience & 3 & 3 \\
        Rotation & [-10.0, 10.0] & [-10.0, 10.0] \\
        Shear & [-10.0, 10.0] & [-10.0, 10.0] \\
        Gaussian noise & $0 \pm 0.5$ & $0 \pm 0.5$ \\
        Contrastive-loss & 0 & - \\
        LoRA rank & - & 8\\
        LoRA alpha & - & 16\\
        LoRA droput & - & 0.3\\
        LoRA bias & - & None \\
        \bottomrule
    \end{tabular}
    \captionsetup{justification=centering}
    \caption{Finetuning hyperparameters}
    \label{tab:finetuning_setup}
\end{table}

\begin{figure}[t!]
\begin{center}
    \includegraphics[width=0.9\columnwidth]{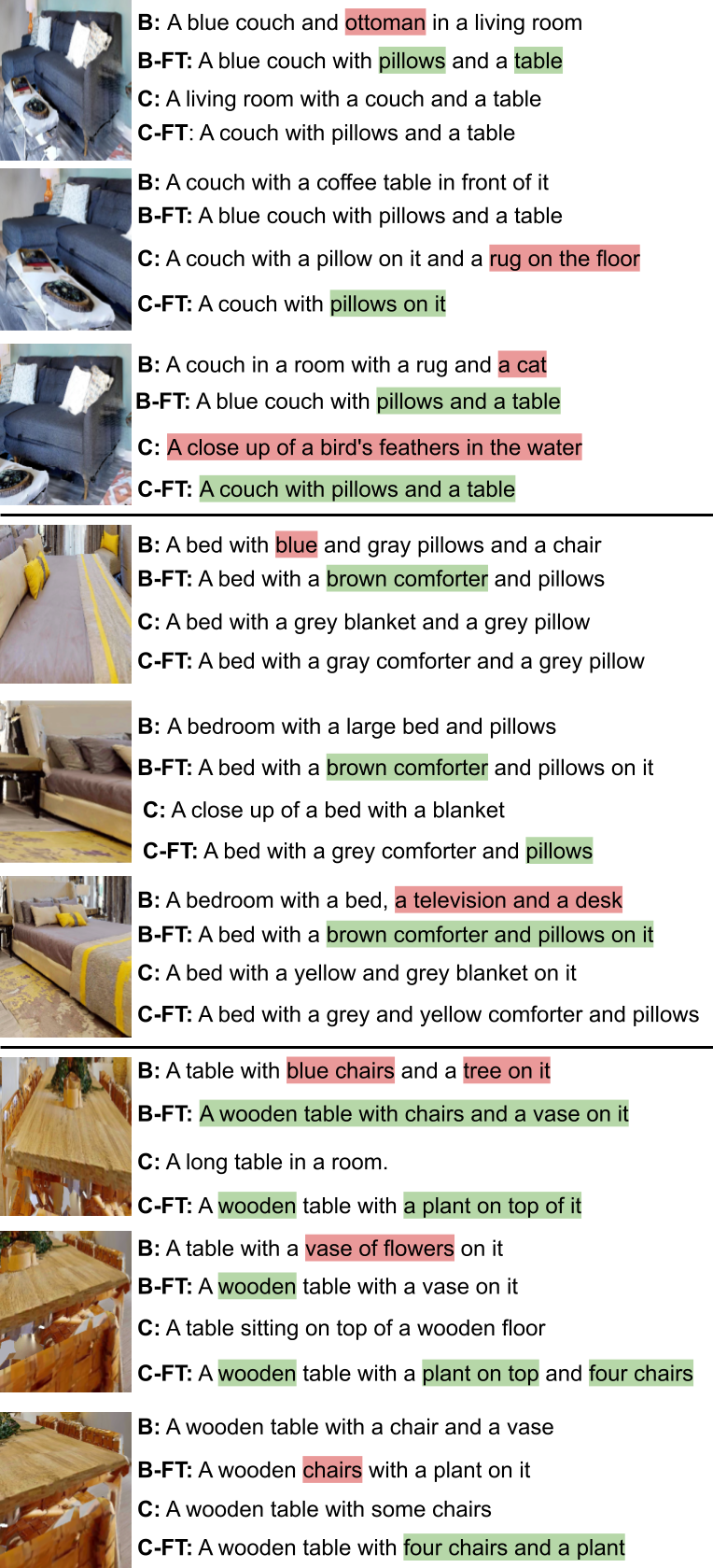}
\end{center}
\caption{Examples of predicted captions before and after fine-tuning for different object views. We highlight \colorbox{BrickRed!60}{mistakes} and \colorbox{YellowGreen!50}{correct details} in generated captions. KEYS -- C: off-the-shelf CoCa~\cite{yu2022coca}, C-FT: fine-tuned (triplet loss) CoCa, B: BLIP2~\cite{li2023blip}, B-FT: fine-tuned (triplet loss) BLIP2. 
}
\label{fig:captioning_qualitative}
\end{figure}

\section{Ablation studies}
\textbf{Pseudo-caption ablation.} Results in Tab.~\ref{tab:llm_captioner_results} show that compared to LLaMa, Mistral achieves -1.85 p.p. in SP and -0.42 p.p. in CI while Qwen achieves -1.41 p.p. in SP and -0.13 p.p. in CI using BLIP2. Also using CoCa, LLaMa outperforms other LLMs. Varying the LLMs, our prompt with caption frequency outperforms ECO and IC3 (Tab.1 main paper).
\begin{table}[t!]
    \centering
    \scriptsize
    \setlength\tabcolsep{1.6pt}
    \renewcommand{\arraystretch}{0.9}
    \begin{tabular}{l l c c c c c c}
    \toprule
    \textbf{Captioner} & \textbf{LLM} & $B_4$ & $M$ & $R_L$ & $CI$ & $SP$ & $CS$ \\
    \midrule
    \multirow{1}{*}[-4pt]{CoCa}   & Mistral-7B    & 17.21 & 22.07 & 44.70 & 1.13 & 29.29 & 71.63 \\
       & Qwen3-8B  & 19.95 & 23.95 & 49.60 & 1.31 & 33.59 & 73.37 \\
    \multirow{1}{*}[-4pt]{BLIP-2} & Mistral-7B    & 16.46 & 24.05 & 48.02 & 1.20 & 33.73 & 72.11 \\
    & Qwen3-8B  & 19.37 & 24.62 & 51.43 & 1.49 & 34.17 & 73.02 \\
    \bottomrule
    \end{tabular}
    \caption{LD-CPS ablation on HM3D with CLA.}
    \vspace{-0.5cm}
    \label{tab:llm_captioner_results}
\end{table}

\textbf{Ablation study on triplet loss.}
Table~\ref{tab:finetuning_ablation_testing_set_performance} shows the effect of the $\lambda_{tr}$ coefficient on the captioners fine-tuning. We experiment $\lambda_{tr} = \{1, 0.5, 0.1\}$. Results show that a choice of $\lambda_{tr}=0.1$ achieves higher performance values compared to standard fien-tuning and other triplet weights. 
With $\lambda_{tr}=1$ the fine-tuned model has worse performance values compared to other triplet loss weights and standard fine-tuned model with all the exploration policies apart from CLA.

\begin{table}[t!] 
    \centering
    \scriptsize
    \setlength\tabcolsep{2pt}
    \resizebox{\columnwidth}{!}{
    \begin{tabular}{r r r r r c c c c c c}
    \toprule
     & \textbf{Captioner} & \textbf{Pseudo-caption} & \textbf{Policy}  & \textbf{Fine-tuning} & $B_4$ & $M$ & $R_L$ & $CI$ & $SP$ & $CS$ \\
    \midrule
    \multirow{13}{*}{\rotatebox[origin=c]{90}{Gibson}} & \multirow{13}{*}{CoCa} & - & - & NoFT & 12.15 & 21.47 & 42.83 & 0.63 & 28.30 & 63.30 \\
    & & \multirow{13}{*}{Ours} & \multirow{4}{*}{Random} & Standard & 19.55 & 23.36 & 51.10 & 1.01 & 33.21 & 70.11  \\
    & & & & $\lambda_{tr} = 1.0$ & 18.94 & 23.73 & 51.95 & 0.97 & 34.11 & 71.18 \\
    & & & & $\lambda_{tr} = 0.5$ & 19.49 & 24.52 & 52.28 & 1.03 & 35.56 & 73.87\\
    & & & & $\lambda_{tr} = 0.1$ & 19.85 & 24.83 & 53.10 & 1.12 & 35.58 & 73.26\\
    \cmidrule{4-11}
    & &  & \multirow{4}{*}{Frontier} & Standard & 17.93 & 24.92 & 51.00 & 1.08 & 35.65 & 73.20 \\
    & & & & $\lambda_{tr} = 1.0$ & 16.53 & 23.58 & 48.97 & 0.89 & 30.22 & 69.28 \\
    & & & & $\lambda_{tr} = 0.5$ & 16.93 & 24.72 & 50.46 & 0.95 & 33.20 & 72.87\\
    & & & & $\lambda_{tr} = 0.1$ & 17.17 & 24.82 & 51.43 & 0.94 & 36.10 & 73.44\\
    \cmidrule{4-11}
    & & & \multirow{4}{*}{CLA} & Standard &  13.70 & 20.50 & 45.43 & 0.72 & 26.48 & 61.76 \\
    & & & & $\lambda_{tr} = 1.0$ & 14.19 & 21.12 & 45.97 & 0.72 & 27.00 & 60.89\\
    & & & & $\lambda_{tr} = 0.5$ & 17.72 & 24.36 & 50.71 & 0.99 & 35.53 & 70.86\\
    & & & & $\lambda_{tr} = 0.1$ & 14.96 & 20.79 & 45.50 & 0.71 & 25.30 & 63.21\\
    \midrule
    \multirow{13}{*}{\rotatebox[origin=c]{90}{HM3D}} & \multirow{13}{*}{CoCa} & - & - & NoFT & 09.93 & 17.36 & 38.91 & 0.44 & 22.19 & 62.08 \\
    & & \multirow{13}{*}{Ours} & \multirow{4}{*}{Random} & Standard & 17.34 & 21.68 & 47.46 & 0.82 & 26.90 & 72.67  \\
    & & & & $\lambda_{tr} = 1.0$ & 16.61 & 19.46 & 44.83 & 0.74 & 25.84 & 70.33 \\
    & & & & $\lambda_{tr} = 0.5$ & 16.51 & 20.92 & 46.29 & 0.81 & 27.41 & 72.53\\
    & & & & $\lambda_{tr} = 0.1$ & 16.57 & 21.39 & 46.73 & 0.83 & 29.36 & 72.77\\
    \cmidrule{4-11}
    & &  & \multirow{4}{*}{Frontier} & Standard & 19.04 & 21.73 & 47.12 & 0.87 & 28.64 & 72.47 \\
    & & & & $\lambda_{tr} = 1.0$ & 18.51 & 21.47 & 46.23 & 0.83 & 27.42 & 70.28 \\
    & & & & $\lambda_{tr} = 0.5$ & 16.80 & 21.42 & 45.44 & 0.74 & 27.43 & 70.32\\
    & & & & $\lambda_{tr} = 0.1$ & 19.11 & 22.05 & 49.02 & 0.95 & 30.21 & 73.99\\
    \cmidrule{4-11}
    & & & \multirow{4}{*}{CLA} & Standard &  16.35 & 20.62 & 46.19 & 0.80 & 28.38 & 70.19 \\
    & & & & $\lambda_{tr} = 1.0$ & 17.01 & 21.33 & 47.01 & 8.75 & 28.71 & 71.02\\
    & & & & $\lambda_{tr} = 0.5$ & 17.31 & 21.55 & 47.19 & 0.79 & 28.74 & 71.74\\
    & & & & $\lambda_{tr} = 0.1$ & 17.57 & 20.73 & 47.81 & 0.81 & 29.36 & 72.23\\
    \bottomrule
    \end{tabular}
    }
    \caption{Comparison of captioning performance on Gibson and HM3D testing sets, varying fine-tuning method and policy. \\ 
    KEYS -- $B_4$: BLEU, $M$: METEOR, $R_L$: ROUGE-L, $CI$: CIDER, $SP$: SPICE, $CS$: cosine similarity between SBERT embedding of prediction and annotation.}
    \label{tab:finetuning_ablation_testing_set_performance}
\end{table}

\section{Inference time} On an NVIDIA Tesla V100 16GB, the time for an exploration step is 0.266 s for CLA equipped with CoCa, 0.271 s equipped with BLIP2 (averaged over 300 steps); of these timings, the captioning inference time is 6.65 ms for CoCa and 10.2 ms for BLIP2 (averaged over 754 bounding boxes); and the pseudo-captioning time per object instance is 1.02 s (averaged over all 160 testing objects).

\section{Qualitative comparison of captioners prediction}
In Fig.~\ref{fig:captioning_qualitative} we present some qualitative examples of predicted captions before and after fine-tuning (with triplet loss) for both CoCa and BLIP2. Off-the-shelf models predict captions containing wrong attributes (colors) or object categories not present in the images, e.g., cat, bird's feathers, television. On the contrary, the fine-tuned models show higher accuracy in describing details of objects and generate predictions that are more consistent across different views of the same object,  showing that learning the unique pseudo-caption for each object instance combined with the triplet loss enhances captions accuracy and consistency. 

\section{Pseudo-captioning and ChatGPT prompts}
\label{sec:prompt}
Listing~\ref{lst:promptchatGPT} shows the text prompt used to generate results with ChatGPT~\cite{openai2025chatgpt}. We use ChatGPT to generate captions from the objects crop.

Listing~\ref{lst:prompt} shows the proposed text prompt for the pseudo-caption generation, where \texttt{str(captions\_freq\_list)} is a list of captions for different views of the same object with their frequency. 
Our approach combines the frequency of the captions with in-context learning (provided examples of how to solve the task), to account for wrong or inconsistent captions.

\begin{lstlisting}[float,basicstyle=\small,caption=The prompt for off-the-shelf ChatGPT.,label=lst:promptchatGPT]
Please provide an image caption for the provided picture. Do not
use more than 5-7 words to describe the object in the image. Use 
simple words and don't use more than one adjective per noun. 
Some examples: 'A red couch with pillows on it','A television
set on top of a table','A grey patterned armchair'
\end{lstlisting}

\begin{lstlisting}[float,basicstyle=\small,caption=The prompt for our proposed method.,label=lst:prompt]
You are an advanced language model tasked with generating a 
concise and accurate caption for an object. You are given a list of 
captions along with their frequencies. Each caption may 
represent a different viewpoint and might not always be accurate. 
Additionally, you are provided with the correct object class to 
describe. Your goal is to generate a single, coherent caption that 
accurately describes the main object, based on the provided 
information. The generated caption should not exceed 20 words 
and must be encapsulated within <Caption> ... </Caption> 
tags.
Here is the format of the input you will receive:
[[frequency, "caption"]]

Example Input:
[[5, "A red apple on a table"], [3, "A shiny red apple"], [1, "A red fruit"], [2, "A red apple"]]
Example Output:
<Caption>A shiny red apple on a table</Caption>
Example Input:
[[8, "A small brown dog"], [3, "A dog"], [4, "A small dog"], [1, "A brown animal"]]
Example Output:
<Caption>A small brown dog</Caption>
Example Input:
[[6, "A blue car parked on the street"], [4, "A car"], [2, "A blue vehicle"], [1, "A car on the street"]]
Example Output:
<Caption>A blue car parked on the street</Caption>
Example Input:
[[7, "A cat sitting on a windowsill"], [5, "A windowsill cat"], [2, "A cat"], [1, "A windowsill"]]
Example Output:
<Caption>A cat sitting on a windowsill</Caption>
Example Input:
[[5, "A wooden table with a plate on it"], [2, "A table with a plate and a couch in the room"], 
[3, "A wooden table"], [1, "A plate on a wooden table"]]
Example Output:
<Caption>A wooden table with a plate on it</Caption>

Your Task:
1. Analyze the provided list of captions and their frequencies.
2. Synthesize an accurate caption that reflects the most reliable and frequent details.
3. Ensure the generated caption describes only the main objects and mentions other objects only in relation to the main object.
4. Ensure the generated caption is no longer than 20 words.
5. Encapsulate your generated caption within <Caption> ... </Caption> tags.

Input:
{str(captions_freq_list)}

Output:
\end{lstlisting}

\section{Performance on occluded objects}
\begin{figure}[t!]
\centering
\includegraphics[width=\linewidth]{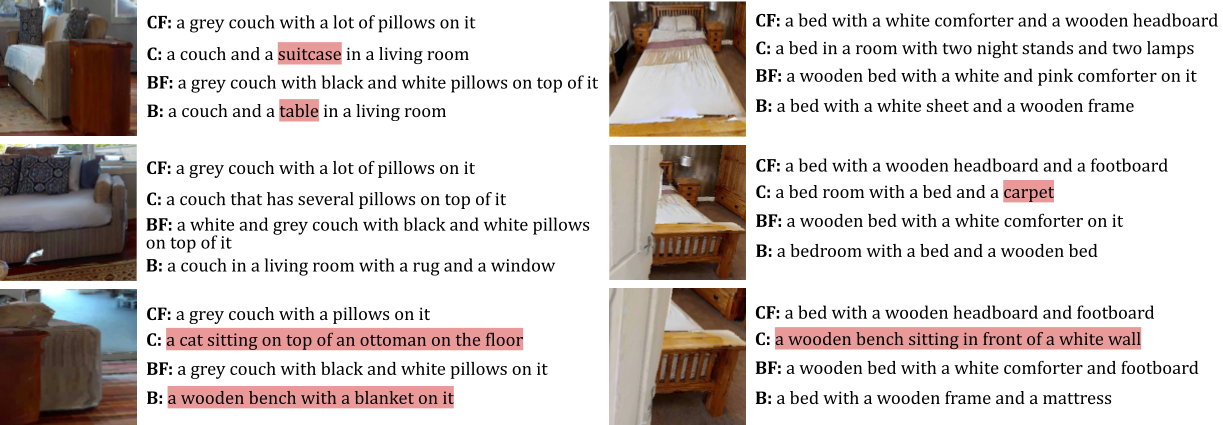}
\caption{Captioning with different levels of occlusion. \textbf{C:} CoCa; \textbf{B:} BLIP2; \textbf{F:} fine-tuned (triplet loss)}
\label{fig:occlusion_performance}
\end{figure}

Tab.~\ref{tab:occlusion_results} reports a comparison between off-the-shelf and fine-tuned captioners (using triplet loss) on a manually collected dataset of 50 images (25 from Gibson and 25 from HM3D) of objects, with 5 images per dataset for each category in the main paper, under varying levels of occlusion based on visual inspection. The performance is evaluated using standard captioning metrics (BLEU-4, METEOR, ROUGE-L, CIDEr, SPICE and cosine similarity), and results are averaged across Gibson and HM3D. The results show that our fine-tuned models outperform off-the-shelf captioners for both CoCa and BLIP-2, indicating improved robustness to partial views and occlusions. Qualitative examples in Fig.~\ref{fig:occlusion_performance} illustrate the increased consistency and accuracy of the generated captions in challenging visual conditions.

\begin{table}[t!]
    \centering
    \scriptsize
    \setlength\tabcolsep{1.6pt}
    \renewcommand{\arraystretch}{0.9}
    \begin{tabular}{l l c c c c c c}
    \toprule 
    \textbf{Env} & \textbf{Finetuning} & $B_4$ & $M$ & $R_L$ & $CI$ & $SP$ & $CS$ \\
    \midrule
    \multirow{2}{*}{CoCa} 
        & NoFT & 16.01 & 22.19 & 46.37 & 0.88 & 31.07 & 68.71 \\
        & Triplet & \textbf{17.95} & \textbf{24.10} & \textbf{48.20} & \textbf{1.05} & \textbf{32.98} & \textbf{74.03} \\
    \midrule
    \multirow{2}{*}{BLIP-2} 
        & NoFT & 20.85 & 23.18 & 50.39 & 1.32 & 36.48 & 73.92 \\
        & Triplet & \textbf{22.70} & \textbf{25.29} & \textbf{53.12} & \textbf{1.38} & \textbf{39.90} & \textbf{77.58} \\
    \bottomrule
    \end{tabular}
    \caption{Comparison of captioning performance on occluded objects. KEYS -- $B_4$: BLEU, $M$: METEOR, $R_L$: ROUGE-L, $CI$: CIDER, $SP$: SPICE, $CS$: cosine similarity between SBERT embedding of predicted and annotated captions.} 
    \label{tab:occlusion_results}
\end{table}

\end{document}